\def\year{2022}\relax
\documentclass[letterpaper]{article} 
\usepackage{aaai22}  
\usepackage{times}  
\usepackage{helvet}  
\usepackage{courier}  
\usepackage[hyphens]{url}  
\usepackage{graphicx} 
\usepackage{natbib}  
\usepackage{caption} 
\usepackage{amsfonts}
\usepackage{amsmath}
\usepackage{multirow}
\usepackage{color}
\usepackage{booktabs}
\DeclareCaptionStyle{ruled}{labelfont=normalfont,labelsep=colon,strut=off} 
\usepackage{algorithm,algpseudocode}

\usepackage{subfigure}

\algnewcommand{\algorithmicforeach}{\textbf{for each}}
\algdef{SE}[FOR]{ForEach}{EndForEach}[1]
{\algorithmicforeach\ #1\ \algorithmicdo}
{\algorithmicend\ \algorithmicforeach}
\usepackage{newfloat}
\usepackage{listings}
\floatstyle{ruled}
\newfloat{listing}{tb}{lst}{}
\floatname{listing}{Listing}
\title{FedCG: Leverage Conditional GAN for Protecting Privacy and Maintaining Competitive Performance in Federated Learning}
\author{
    Yuezhou Wu$^{1,2}$\equalcontrib,
    Yan Kang$^{1}$\equalcontrib \thanks{Corresponding author},
    Jiahuan Luo$^{1}$\equalcontrib,
    Yuanqin He$^{1}$,
    Qiang Yang$^{1,3}$
}
\affiliations{
$^1$AI Group, WeBank, Shenzhen, China\\
$^2$School of Computer Science and Engineering, Sun Yat-sen University, Guangzhou, China \\
$^3$Department of CSE, Hong Kong University of Science and Technology, Hong Kong, China\\
    
\{majorwu, yangkang, jiahuanluo, yuanqinhe, lixinFan\}@webank.com, qyang@cse.ust.hk

}

\usepackage{bibentry}

\begin{document}

\maketitle

\begin{abstract}
Federated learning (FL) aims to protect data privacy by enabling clients to build machine learning models collaboratively without sharing their private data. Recent works demonstrate that information exchanged during FL is subject to gradient-based privacy attacks and, consequently, a variety of privacy-preserving methods have been adopted to thwart such attacks. However, these defensive methods either introduce orders of magnitudes more computational and communication overheads (e.g., with homomorphic encryption) or incur substantial model performance losses in terms of prediction accuracy (e.g., with differential privacy). In this work, we propose \textsc{FedCG}, a novel \underline{fed}erated learning method that leverages \underline{c}onditional \underline{g}enerative adversarial networks to achieve high-level privacy protection while still maintaining competitive model performance. \textsc{FedCG} decomposes each client's local network into a private extractor and a public classifier and keeps the extractor local to protect privacy. Instead of exposing extractors, \textsc{FedCG} shares clients' generators with the server for aggregating clients' shared knowledge aiming to enhance the performance of each client's local networks. Extensive experiments demonstrate that \textsc{FedCG} can achieve competitive model performance compared with FL baselines, and privacy analysis shows that \textsc{FedCG} has a high-level privacy-preserving capability.

\end{abstract}

\section{Introduction}
Deep neural networks (DNN) have achieved dramatic success in many areas, including computer vision, natural language processing, and recommendation systems. Their success largely depends upon the availability of a wealthy amount of training data. In many real-world applications, however, training data is typically distributed across different organizations, which are unwilling to share their data because of privacy and regulation concerns directly. To alleviate these concerns, federated learning (FL) \cite{mcmahan2017communication} is proposed to enable multiple clients to collaboratively build DNN models without sharing clients' private data. 

Despite the privacy-preserving capability introduced by FL, recent works have empirically demonstrated that the classic federated averaging method (FedAvg \cite{mcmahan2017communication}) and its variants (e.g., FedProx \cite{li2018federated}) are vulnerable to gradient-based privacy attacks such as the deep leakage from gradients (DLG) \cite{zhu2020deep}, which is able to reconstruct the original data of clients from publicly shared gradients and parameters. Varieties of technologies have been leveraged to further improve the privacy of FL, the most popular ones are homomorphic encryption (HE) \cite{gentry2009fully,aono2017privacy} and differential privacy (DP) \cite{dwork2014algorithmic}. HE provides a high-level security guarantee by encrypting exchanged information among clients. Nonetheless, its extremely high computation and communication cost make it unsuitable to DNN models that typically consist of numerous parameters. While DP imposes a low complexity on FL, it causes precision loss and still suffers from data recovery attacks. To prevent data leakage and still enjoy the benefits of FL, \textsc{FedSplit} \cite{gupta2018distributed, gu2021federated} combining split learning~\cite{vepakomma2018split} and federated learning proposes to split a client's network into private and public models,  and protect privacy by hiding private model from the server. However, \textsc{FedSplit} experiences a non-negligible performance drop.

In this work, we propose \textsc{FedCG}, a novel \underline{fed}erated learning method that leverages \underline{c}onditional \underline{g}enerative adversarial networks~\cite{mirza2014conditional} to achieve high-level privacy protection resisting DLG attack while still maintaining competitive model performance compared with baseline FL methods. More specifically, \textsc{FedCG} decomposes each client's local network into a \textit{private} extractor and a \textit{public} classifier, and keeps the extractor local to protect privacy. The novel part of \textsc{FedCG} is that it shares clients' generators in the place of extractors with the server for aggregating clients' shared knowledge aiming to enhance model performance. This strategy has two immediate advantages. First, the possibility of clients' data leakage is significantly reduced because no model that directly contacts with original data is exposed, as compared to FL methods in which the server has full access to clients' local networks (e.g., \textsc{FedAvg} and \textsc{FedProx}). Second, the server can aggregate clients' generators and classifiers using knowledge distillation~\cite{hinton2015distilling} without accessing any public data. 

The main contributions of this work are:
\begin{itemize}
\item To the best of our knowledge, this work is the first attempt to integrate cGAN into FL aiming to protect clients' data privacy by hiding local extractors from the server while enabling clients to have competitive model performance by integrating shared knowledge embedded in the global generator. 

\item 

Extensive experiments demonstrate that \textsc{FedCG} can achieve competitive performance compared with varieties of baseline FL methods. The privacy analysis proves that \textsc{FedCG} has a high-level privacy-preserving capability against gradient inversion attacks.

\end{itemize}

\section{Related Work}

\subsection{Federated Learning}
Federated learning (FL) is a distributed machine learning paradigm that enables clients (devices or organizations) to train a machine learning model collaboratively without exposing clients' local data. The concept of FL was first proposed by \cite{mcmahan2017communication}. Further, many serious challenges emerge with the development of FL. Particularly, the "naked" FL methods without any privacy protection mechanism are proven to be vulnerable to data recovery attacks such as deep leakage ~\cite{zhu2020deep,zhao2020idlg,geiping2020inverting} and model inversion~\cite{fredrikson2015model}. Therefore, a wealth of technologies have been proposed to improve the privacy of FL. The most popular ones are homomorphic encryption (HE) and differential privacy (DP). However, HE is extremely computationally expensive, while DP suffers from non-negligible precision loss. Another school of FL methods~\cite{gupta2018distributed,poirot2019split,gu2021federated} tries to strike a balance between privacy and efficiency by splitting a neural network into private and public models and sharing only the public one.

\subsection{DLG in Federated Learning}
Federated learning is proposed to protect data privacy by keeping private data localized and sharing only model gradients or parameters. However, recent research on \textit{Deep Leakage from Gradients} (DLG) demonstrates~\cite{zhu2020deep} that shared gradients can actually leak private training data. Particularly, DLG can achieve pixel-wise level data recovery without any assistance information. A follow-up work~\cite{zhao2020idlg} shows that label information can also be restored from gradients of the last layer of a client's model. \cite{geiping2020inverting} further extends DLG to more realistic settings where gradients are averaged over several iterations or several images and shows that users' privacy is not protected in these settings. GRNN~\cite{ren2021grnn} improves DLG by recovering fake data and labels through a generative model instead of regressing them directly from random initialization. 

\subsection{GAN In Federated Learning} 
Recent research works that utilize GAN in the FL setting focus mainly on two lines of works: One is leveraging GAN to perform data recovery attacks. \cite{hitaj2017deep} assumes a malicious client that utilizes the shared model as the discriminator to train the generator in a GAN. Then, the trained generator is used to mimic the training samples of the victim client. \cite{wang2019beyond} assumes the server is malicious. The malicious server first learns representations of the victim client's data through DLG and then leverages these representations to train the generator, which eventually can generate the victim's private data. Another is to train high-quality GAN across distributed data sources under privacy, efficiency, or heterogeneity constraints. MD-GAN~\cite{hardy2019md} proposes a system that the server hosts the generator while each client hosts a discriminator. The discriminators communicate with each other in a peer-to-peer fashion for improving computational efficiency. FedGAN~\cite{rasouli2020fedgan} trains a GAN across Non-IID data sources in a communication efficient way, but it may produce biased data. The follow-up work~\cite{mugunthan2021bias} is able to generate bias-free synthetic datasets using FedGAN by fine-tuning the federated GAN with synthesized metadata. Fed-TGAN~\cite{zhao2021fed} is proposed to learn complex tabular GAN on non-identical clients. 

\section{Proposed Method}

\subsection{Problem Formulation}

In this work, we consider typical federated learning (FL) setting that includes a central server and $N$ clients holding private datasets $\{\mathcal{X}_1, \mathcal{X}_2,...,\mathcal{X}_n\}$. These private datasets share the same feature space but have different sample spaces. 

\begin{figure}[ht!]
  \centering
  \includegraphics[width=.45\textwidth]{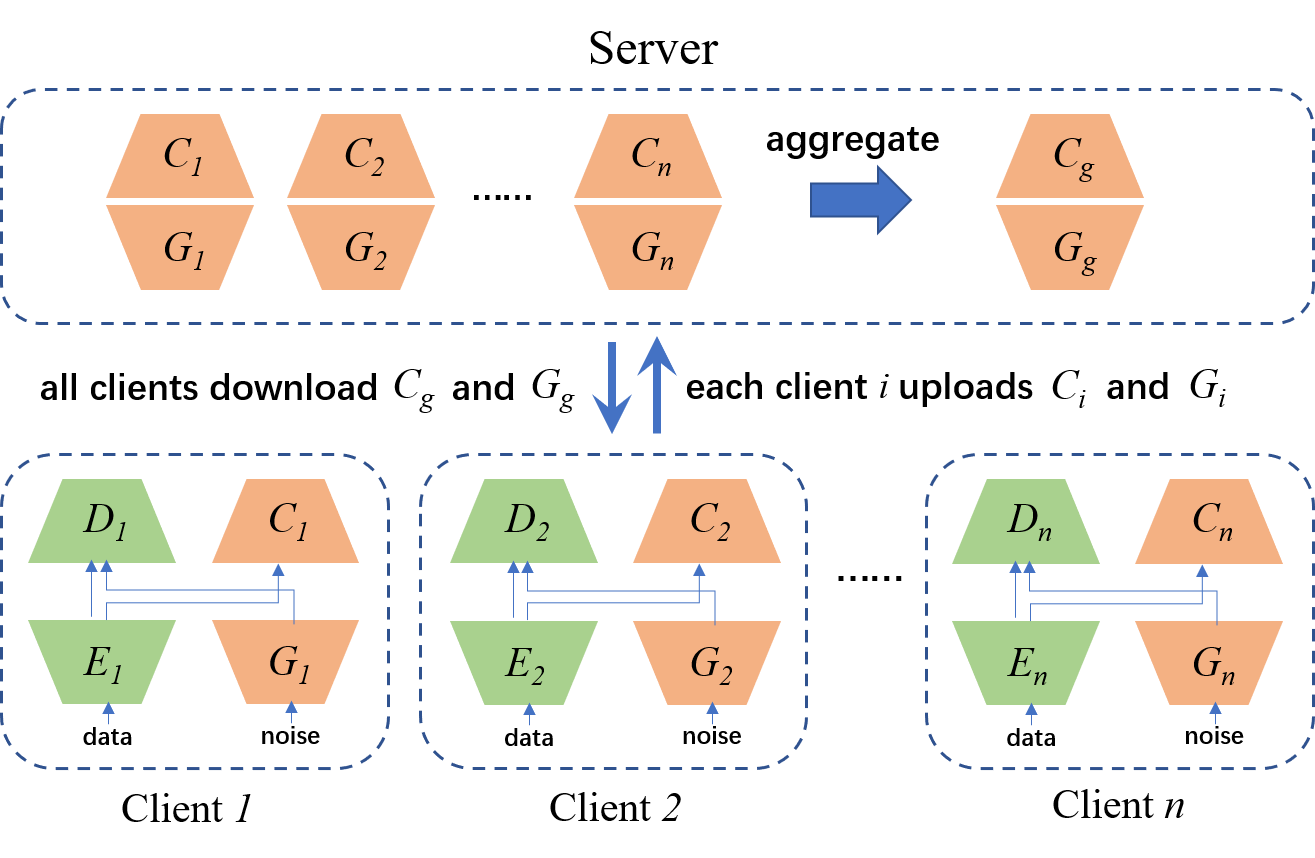}
  \caption{\textbf{Overview of FedCG}. 
  }
  \label{framework}
\end{figure}

Each client $i$ has a classification network parameterized by $\theta_{E_i, C_i}:=[\theta_{E_i};\theta_{C_i}]$ consists of a feature extractor $E_i:\mathcal{X}_i\rightarrow \mathbb{R}^{d}$ parameterized by $\theta_{E_i}$, and a classifier $C_i:\mathbb{R}^d\rightarrow \mathbb{R}^{c}$ parameterized by $\theta_{C_i}$, where $d$ is the feature dimension and $c$ is the number of classes. For protecting privacy and maintaining competitive performance, each client is provided with a conditional GAN (cGAN) consisting of a generator $G_i:\mathcal{Z}\rightarrow \mathbb{R}^{d}$ parameterized by $\theta_{G_i}$, and a discriminator $D_i:\mathbb{R}^{d} \rightarrow \mathcal{I}$ parameterized by $\theta_{D_i}$, where $\mathcal{Z}$ is the Gaussian distribution and $\mathcal{I}$ 
indicates a single scalar in the range of [0, 1]. The training procedure of the cGAN is performed locally aiming to train the generator $G_i$ to approximate the extractor $E_i$ such that $G_i(z, y)$ captures the distribution of features extracted by $E_i(x|y))$.

As illustrated in Figure \ref{framework}, the workflow of \textsc{FedCG} goes as follows: in each FL communication round, each client $i$ uploads its $G_i$ and $C_i$ to the server once the local training is completed while keeps the $E_i$ and $D_i$ local to strengthen privacy protection. Then, the server applies knowledge distillation to build a global generator $G_g$ and a global classifier $C_g$. Next, clients download $G_g$ and $C_g$ to replace their corresponding local models and start the next training iteration.

In our \textsc{FedCG}, clients collaboratively train the global generator and classifier with the help of the central server, while each client leverages the global generator and global classifier to build a personalized local classification network that can perform well on its local test dataset. We will elaborate on this in the following two sections. 

\subsection{Two-stage Client Update}

\begin{figure}[htbp]
  \centering
  \includegraphics[width=.45\textwidth]{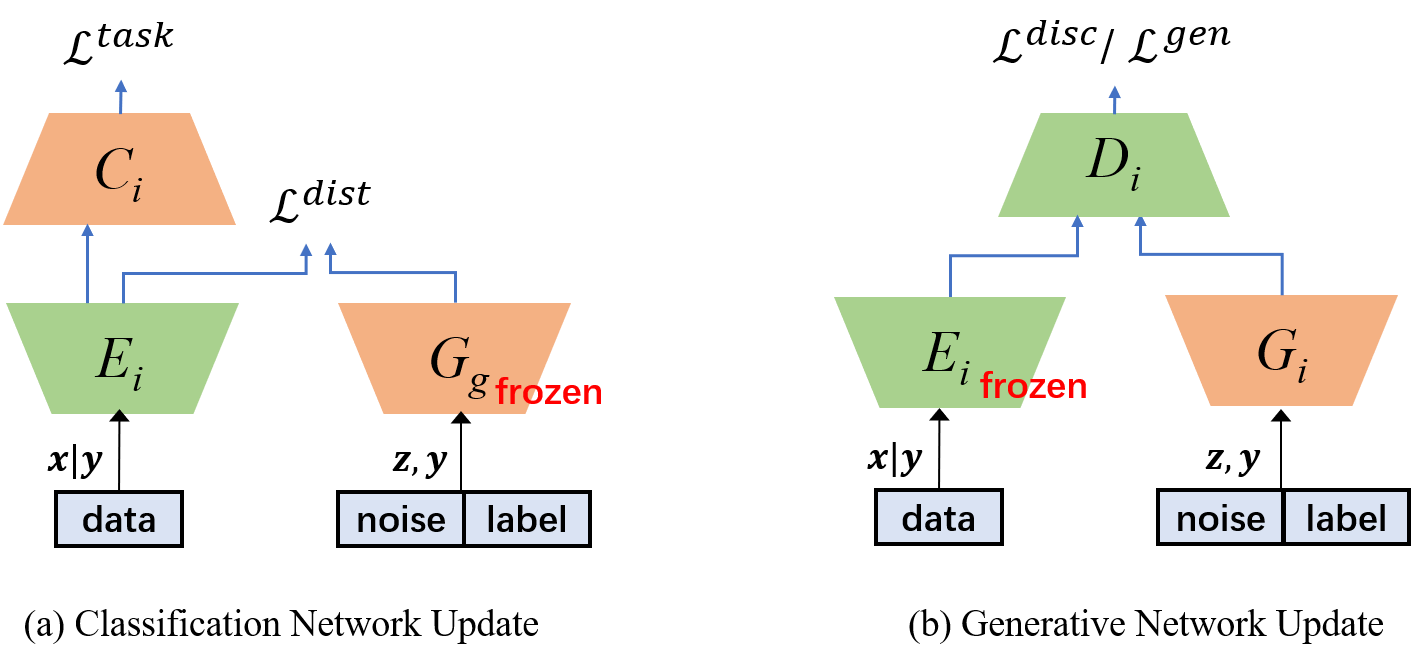}
  \caption{\textbf{Two-stage client update}. (a) Classification network update. (b) Generative network update. 
  } 
  \label{client_train}
\end{figure}

The client's local training procedure involves two stages: classification network update and generative network update. Figure \ref{client_train} illustrates the two stages while Algorithm \ref{client_update_alg} describes the detailed procedure.

In the classification network update stage (Algo \ref{client_update_alg}, lines 3-9), each client $i$ optimizes its extractor $\theta_{E_i}$ and classifier $\theta_{C_i}$ by minimizing the classification loss: 
\begin{equation}
\label{equ:classification}
\mathcal{L}^{cls} = \mathbb{E}_{x,y \sim \mathcal{X}_i}\ \Omega \big( C_i(E_i(x|y;\theta_{E_i});\theta_{C_i}), y \big),
\end{equation}where $\Omega$ is cross-entropy function. In addition, client $i$ also wants to integrate the shared knowledge embedded in the global generator (aggregated in the previous round) into its local extractor. To this end, it freezes the global generator $\theta_{G_g}$ and optimizes its local extractor $\theta_{E_i}$ by minimizing the mean-square-error loss as follows: 
\begin{equation}
\mathcal{L}^{mse} = \mathbb{E}_{x,y \sim \mathcal{X}_i}\mathbb{E}_{z \sim \mathcal{Z}} || E_i(x|y;\theta_{E_i})-G_g(z,y;\theta_{G_g}) || ^2.
\end{equation}

Then, we have the task loss $\mathcal{L}^{task}$, which is the combination of $\mathcal{L}^{cls}$ and $\mathcal{L}^{mse}$, and is formalized by: 
\begin{equation}
\label{equ:class}
\mathcal{L}^{task} = \mathcal{L}^{cls} + \gamma\ \mathcal{L}^{mse},
\end{equation}where the $\gamma$ is a non-negative hyperparameter that adjusts the balance between the two loss terms. In this work, $\gamma$ gradually increases from 0 to 1 as the global generator becomes more accurate in producing feature representations during training. 

\begin{algorithm}[!ht] 
\caption{Two-stage Client Update}
\label{client_update_alg}
\begin{algorithmic}[1]

\Statex \textbf{Input:} 
clients' datasets $\{\mathcal{X}_i\}_{i=1}^n$;
clients' extractors, classifiers, generators and discriminators: $\{E_i(\cdot;\theta_{E_i}),  C_i(\cdot;\theta_{C_i}), G_i(\cdot;\theta_{G_i}), D_i(\cdot;\theta_{D_i})\}_{i=1}^n$; 
global generator $\{G_g(\cdot;\theta_{G_g})$ and global classifier $C_g(\cdot;\theta_{C_g}\})$; learning rate $\eta_1$, $\eta_2$; local training epoch $T$ 
 
\State Clients receive $\theta_{G_g}$ and $\theta_{C_g}$ from the server.
\ForEach{client $i = 1,...,N$ in parallel}
\State $\theta_{C_i} \leftarrow \theta_{C_g}$;
\For {$t \in \{1,...,T\}$} 
    \ForAll {$x,y \in \mathcal{X}_i$}
    \State sample $z$ from $N(0,1)$
    \State $\theta_{E_i,C_i} \leftarrow \theta_{E_i,C_i} - \eta_1 \nabla _{\theta_{E_i,C_i}} \mathcal{L}^{task}(x,y,z) $
    \EndFor
\EndFor
\State $\theta_{G_i} \leftarrow \theta_{G_g}$;
\For {$t \in \{1,...,T\}$} 
\ForAll {$x,y \in \mathcal{X}_i$}
\State sample $z$ from $N(0,1)$
\State $\theta_{D_i} \leftarrow \theta_{D_i} - \eta_2 \nabla _{\theta_{D_i}} \mathcal{L}^{disc}(x,y,z)$ 
\State $\theta_{G_i} \leftarrow \theta_{G_i} - \eta_2 \nabla _{\theta_{G_i}} \mathcal{L}^{gen}(y,z)$ 
\EndFor
\EndFor
\EndForEach
\State Client $i$ sends $\theta_{G_i}$ and $\theta_{C_i}$ to server.
\end{algorithmic}
\end{algorithm}

In the generative network update stage (Algo \ref{client_update_alg}, lines 10-18), each client $i$ wants to approximate its local generator's output to the feature representations extracted by its local extractor. To this end, it freezes the parameters $\theta_{E_i}$ of extractor $E_i$ and conducts a cGAN training procedure to train the generator. More specifically, it samples a mini-batch of training data $(x, y)$ and feeds $x$ to the $E_i$ to obtain the "ground-truth" feature representations $h$. Then, it randomly generates Gaussian noises $z$ with the same batch size and feeds $(z, y)$ to generator $G_i$ to generate estimated feature representations $\hat{h}$. Next, it feeds $h$ and $\hat{h}$ to discriminator $D_i$ to calculate discriminator loss $\mathcal{L}^{disc}$ and generator loss $\mathcal{L}^{gen}$ according to (\ref{equ:gan}), and alternatively minimize the two losses to optimize the generator $\theta_{G_i}$ and discriminator $\theta_{D_i}$.  
\begin{equation}
\label{equ:gan}
\begin{split}
\mathcal{L}^{disc} = \ \mathbb{E}_{x,y \sim \mathcal{X}_i}\mathbb{E}_{z \sim \mathcal{Z}}  \big [& \log{(1-D_i(E_i(x|y;\theta_{E_i});\theta_{D_i}))} \\ + & \log{D_i(G_i(z,y;\theta_{G_i});\theta_{D_i})} \big], \\
\mathcal{L}^{gen} = \ \mathbb{E}_{x,y \sim \mathcal{X}_i}\mathbb{E}_{z \sim \mathcal{Z}} & \log{(1-D_i(G_i(z,y;\theta_{G_i});\theta_{D_i}))}.
\end{split}
\end{equation}

Once the local training is completed, each client $i$ sends its generator $\theta_{G_i}$ and classifier $\theta_{C_i}$ to the server for aggregation.

\subsection{Server Aggregation}

\begin{figure}[ht!]
  \centering
  \includegraphics[width=.43\textwidth]{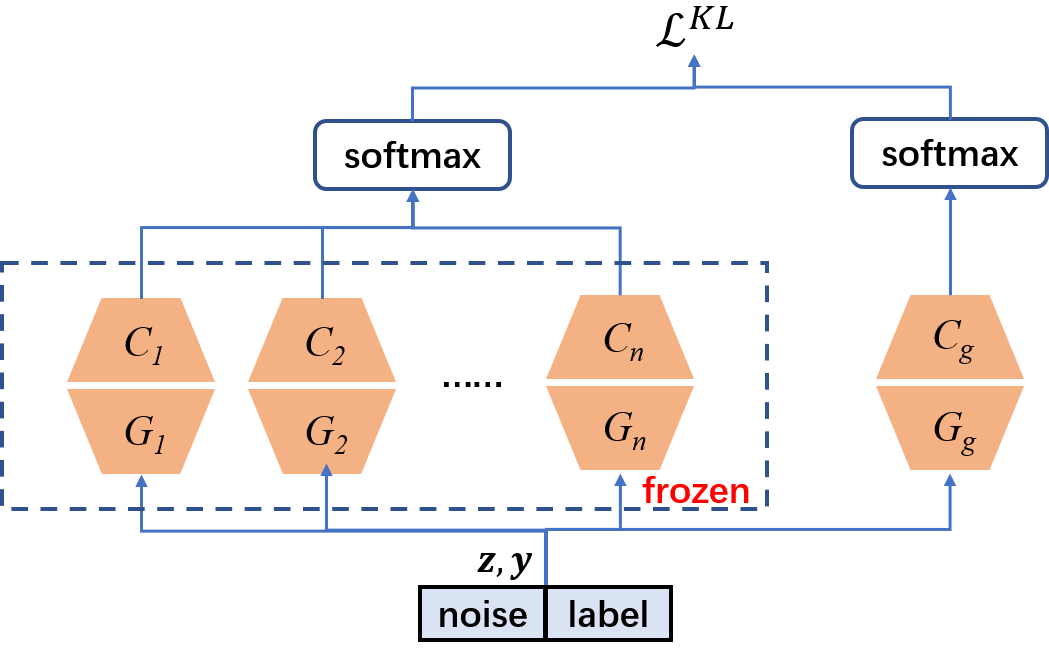}
  \caption{\textbf{Server aggregation.} The server generates Gaussian noise $z$ and class label $y$ as the inputs of clients' $\{G_i\}_{i=1}^n$ and global $G_g$, and it optimizes $\theta_{C_g}$ and $\theta_{C_g}$ by minimizing the KL divergence between the distribution ensembled from $\{C_i\}_{i=1}^n$ and the one outputted from $C_g$.} 
  \label{server_train}
\end{figure}

The server utilizes knowledge distillation (KD)~\cite{lin2020ensemble} to perform the aggregation. One major advantage of $\textsc{FedCG}$ over existing KD methods in FL is that $\textsc{FedCG}$ does not require the server to access any public data in order to perform distillation. 

Figure \ref{server_train} illustrates the server aggregation while Algorithm \ref{server_update_alg} describes the detailed procedure. When the server receives generators $\{G_i\}_{i=1}^n$ and classifiers $\{C_i\}_{i=1}^n$ uploaded by clients, it initializes parameters of the global generator $\theta_{G_g}$ and global classifier $\theta_{C_g}$ by weighted averaging $\{\theta_{G_i}\}_{i=1}^n$ and $\{\theta_{C_i}\}_{i=1}^n$. For distillation (Algo \ref{server_update_alg}, line 3-6), the server first generates a mini-batch of training data $(z, y)$, where labels $y$ are sampled from uniform distribution $\mathcal{U}(0, c)$ and noises $z$ are sampled from Gaussian distribution $N(0, 1)$. Then, according to (\ref{equ:server_distr}) it feeds $(z, y)$ to all generators and calculates two class probability distributions $\mathcal{P}_{c}(y,z)$ and $\mathcal{P}_{s}(y,z)$, the former is ensembled from clients' classifiers while the latter is from the global classifier. Next, the server optimizes the global classifier $\theta_{C_g}$ and generator $\theta_{G_g}$ by minimizing the $\text{KL}$ divergence between the two distributions, according to (\ref{equ:server_kl}). 

\begin{equation}
\label{equ:server_distr}
\begin{split}
\mathcal{P}_{c}(y,z)  &=   \sigma \big ( \sum_{i=1}^n \frac{|\mathcal{X}_i|}{\sum_{i=1}^n|\mathcal{X}_i|} C_i(G_i(y,z;\theta_{G_i});\theta_{C_i}) \big),\\
\mathcal{P}_{s}(y,z) &=  \sigma \big(C_g(G_g(y,z;\theta_{G_g});\theta_{C_g}) \big),\\
\end{split}
\end{equation}

\begin{equation}
\label{equ:server_kl}
\begin{split}
\mathcal{L}^{\text{KL}} &= \mathbb{E}_{y \sim \mathcal{U}}\mathbb{E}_{z \sim \mathcal{Z}} \text{KL}(\mathcal{P}_{c}(y,z), \mathcal{P}_{s}(y,z)),
\end{split}
\end{equation}
where $|\cdot|$ denotes the size of data set, $\text{KL}$ indicates Kullback–Leibler divergence and $\sigma$ is the softmax function. Once the server aggregation is completed, the server dispatches the global generator $\theta_{G_g}$ and global classifier $\theta_{C_g}$ to all clients.

\begin{algorithm}[!ht] 
\caption{Server Aggregation}
\label{server_update_alg}
\begin{algorithmic}[1]
\Statex \textbf{Input:} size of data set $|\mathcal{X}_i|_{i=1}^n$; clients' generators and classifiers ${\{G_i(\cdot;\theta_{G_i}),C_i(\cdot;\theta_{C_i})\}}_{i=1}^n$ ; learning rate $\eta_3$; training iteration $T$, sample batch size $B$.
\State Server receives ${\{\theta_{G_i}\}}_{i=1}^n$ and ${\{\theta_{C_i}\}}_{i=1}^n$ from all clients.

\State $\{\theta_{G_g},\theta_{C_g}\} = \sum_{i=1}^n \frac{|\mathcal{X}_i|}{\sum_{i=1}^n |\mathcal{X}_i|}\{\theta_{G_i}, \theta_{C_i}\}$ 

\For{$t \in \{1,...,T\}$}
\State Sample $(z, y)$, where $z \sim N(0,1)$ and $y \sim \mathcal{U}(0,c)$.
\State $ \{ \theta_{G_g}, \theta_{C_g} \} \leftarrow \{ \theta_{G_g}, \theta_{C_g} \} - \eta_3 \nabla _{\theta_{G_g, C_g}} \mathcal{L}^{\text{KL}}(y,z) $ 
\EndFor

\State Server sends $\theta_{G_g}$ and $\theta_{C_g}$ to all clients.
\end{algorithmic}
\end{algorithm}

\subsection{Privacy Analysis}

In the literature, there are a variety of data recovery methods, among which DLG~\cite{zhu2020deep} is able to achieve exact pixel-wise level data revealing. In this work, we consider the server malicious, and it utilizes DLG to recover original data from a victim client. We evaluate the privacy-preserving capability of \textsc{FedCG} by comparing the quality of image data recovered in \textsc{FedCG}, \textsc{FedAvg} and \textsc{FedSplit}. \textsc{FedAvg} is the representative of FL methods that share the full classification network while \textsc{FedSplit} represents FL methods that share only the public classifier.  (\ref{equ:server_dlg}) shows the DLG loss $\mathcal{L}^{dlg}$ for the malicious server recovering the real data of victim client $i$.
\begin{equation}
\label{equ:server_dlg}
\mathcal{L}^{dlg} = ||\nabla _\theta \mathcal{L}^{cls}(x_i) - \nabla _\theta \mathcal{L}^{cls}(\tilde{x}) ||^2 ,
\end{equation}
where $x_i$ is the real data of victim client $i$ while $\tilde{x}$ is the variable to be trained to approximate $x_i$ by minimizing DLG loss $\mathcal{L}^{dlg}$, which is the distance between $\nabla_\theta \mathcal{L}^{cls}(x_i)$ and $\nabla_\theta \mathcal{L}^{cls}(\tilde{x})$. The former is observed gradients of $\mathcal{L}^{cls}$ (see (\ref{equ:classification})) w.r.t. model parameters $\theta$ for the real data $x_i$, while the latter is estimated gradients for $\tilde{x}$. For \textsc{FedAvg}, the server knows the full network of client $i$, thus $\theta:=\theta_{E_i,C_i}$. While for \textsc{FedSplit}, the server only knows the classifier, and thus $\theta:=\theta_{C_i}$. Although subsequent research works on DLG generally employs cosine similarity as the loss function, the image reconstruction quality of MSE is more satisfactory for LeNet networks \cite{geiping2020inverting}.


Similar to \textsc{FedSplit}, \textsc{FedCG} hides private extractors from the server for protecting privacy. Nonetheless, \textsc{FedCG} shares clients' generators with the server for aggregating shared knowledge. Thus, \textsc{FedCG} has auxiliary information on extractors' output distribution estimated by generators. We define the DLG loss for \textsc{FedCG} as follows:

\begin{equation}
\begin{split}
\label{equ:fedgd_dlg}
\mathcal{L}^{dlg}_{\textsc{FedCG}} = || \nabla _{\theta_{C_i}}\mathcal{L}^{cls}(x_i) - \nabla _{\theta_{C_i}}\mathcal{L}^{cls}(\tilde{x}) ||^2 + \alpha \sum_c \\ (||mean(\tilde{E}(\tilde{x})^c) - \mu^c||^2 + ||var(\tilde{E}(\tilde{x})^c) - \sigma^c||^2),
\end{split}
\end{equation}where $\tilde{E}$ is the estimated extractor whose parameters are not known by the server, and thus it is randomly initialized. The second optimization term of $\mathcal{L}^{dlg}_{\textsc{FedCG}}$ aligns the statistical information of features between the estimated extractor $\tilde{E}$ of the server and the shared generator of victim client $G_i$. $\mu^c$ and $\sigma^c$ are the mean and standard deviation on each channel $c$ of features generated by $G_i$. We will quantitatively measure privacy-preserving capabilities of \textsc{FedAvg}, \textsc{FedSplit} and \textsc{FedCG} according to (\ref{equ:server_dlg}) and (\ref{equ:fedgd_dlg}) in the next section.  

\section{Experiments}

In this section, we compare the performance of our proposed \textsc{FedCG} against FL baselines and evaluate the privacy-preserving capability of \textsc{FedCG}. 

\begin{figure*}[htbp]
    \centering
    \includegraphics[width=7.0in,height=3.8cm]{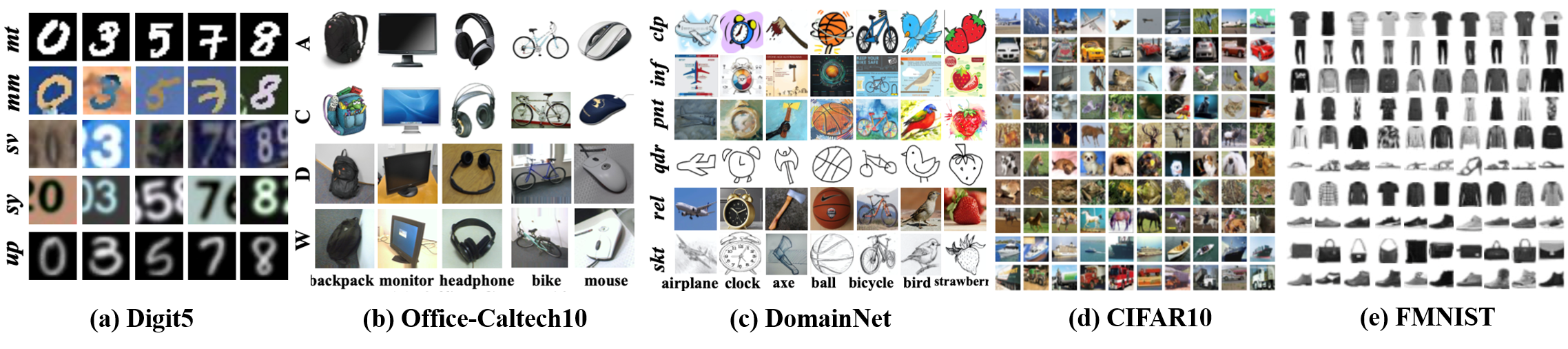}
    \caption{We demonstrate the competitive performance of \textsc{FedCG} on 5 datasets: Digit5 is a collection of 5 benchmarks for digit recognition, namely \textit{MNIST}, \textit{Synthetic Digits}, \textit{MNIST-M}, \textit{SVHN}, and \textit{USPS}. Office-Caltech10 contains 10 Office Supplies from 4 domains: \textit{Amazon}, \textit{DSLR}, \textit{Webcam}, and \textit{Caltech}. DomainNet comprises of 6 domains: \textit{Painting}, \textit{Clipart}, \textit{Infograph}, \textit{Quickdraw}, \textit{Real} and \textit{Sketch}. FMNIST and CIFAR10 are widely used image datasets.}
    \label{datasets_fig}
\end{figure*}

\subsection{Experiment Settings}

\subsubsection{Model Architecture}
LeNet5 \cite{lecun2015lenet} is used as the backbone network for image classification tasks in FL system. The first 2 convolutional layers of LeNet are regarded as \textit{private} extractor, while the latter 3 fully connected layers are regarded as \textit{public} classifier. The generative adversarial network architecture is a modified version of DCGAN \cite{radford2015unsupervised}, in which the size and stride of convolution kernels are adjusted to match the output dimensions of the extractor and generator.

\subsubsection{Datasets} We evaluate model performance of clients on 5 image datasets: FMNIST~\cite{xiao2017fashion}, CIFAR10~\cite{krizhevsky2009learning}, Digit5~\cite{peng2019moment}, Office-Caltech10~\cite{gong2012geodesic}, and DomainNet~\cite{peng2019moment}. Figure \ref{datasets_fig} shows some samples of the 5 datasets. FMNIST and CIFAR10 simulate the independent identically distributed (IID) setting. Digit5, Office-Caltech10, and DomainNet comprise data from multiple domains, and thus they naturally form the Non-IID setting. DomainNet has images across 345 categories, from which we pick the top 10 categories with the most samples for experiments. To keep the image resolution consistent across all experiments, we resize all images in all datasets to 32 $\times$ 32.

\subsubsection{Baselines}
We chose 5 FL baselines from two categories of FL methods. The first category includes \textsc{FedAvg}~\cite{mcmahan2017communication}, \textsc{FedProx} \cite{li2018federated} and \textsc{FedDF} \cite{lin2020ensemble}, in which clients share their full networks with the server. The second one includes \textsc{FedSplit}~\cite{gu2021federated} and \textsc{FedGen}~\cite{zhu2021data}, in which clients share only their public classifiers. More specifically, \textsc{FedAvg} is the most widely used FL method. \textsc{FedProx} introduces a proximal term in the local objective to regularize the local model training. \textsc{FedDF} utilizes knowledge distillation to aggregate local models on the server leveraging unlabeled public data. \textsc{FedSplit} shares only the public classifier of the local classification network to protect privacy. \textsc{FedGen} employs knowledge distillation to train a global generator, which in turn helps clients train local models. In this work, we implement \textsc{FedGen} based on \textsc{FedSplit}, in which only the public classifiers of local networks are shared. We also consider the local classification network trained solely based on local data as a baseline and denote it as \textsc{Local}.

\subsubsection{Configurations}
We perform 100 global communication rounds and 20 local epochs, and each local epoch adopts a mini-batch of 16. All experiments use the Adam optimizer with a learning rate of 3e-4 and a weight decay of 1e-4. For \textsc{FedProx}, we tried proximal term factor in the range of \{0.001, 0.01, 0.1, 1\} and picked the best one. \textsc{FedGen}, \textsc{FedDF} and \textsc{FedCG} perform 2000 iterations in the server for model fusion with a batch size of 16. 

We consider each domain as a client for Digit5, Office-Caltech10 and DomainNet except that \textit{MNIST} and \textit{Painting} are held out as distillation data for 
\textsc{FedDF}. We consider 4- and 8-client scenarios for CIFAR10 and consider a 4-client scenario for FMNIST. For these two datasets, 32000 training samples are held out as distillation data for \textsc{FedDF}. For FMNIST, CIFAR10, and Digit5, we randomly sample 2000 images for each client as the local training set. For Office-Caltech10 and DomainNet, 50\% of the original data of each domain is used as the local training set. We use validation and test datasets on clients to report the best test accuracy over 5 different random seeds. We also leverage \textit{diversity} loss from~\cite{MSGAN} to improve the stability of the generator and adopt early stopping to avoid overfitting. 


\subsection{Experiment Results}

\begin{table*}[ht]
\centering
    \begin{tabular}{l||c|c|c||c|c|c}
    \hline
    \multicolumn{1}{c}{} & \multicolumn{3}{c}{IID} & \multicolumn{3}{c}{non-IID} \\
    \multicolumn{1}{l}{Method} & \multicolumn{1}{c}{FMNIST (4)} & \multicolumn{1}{c}{CIFAR10 (4)} & \multicolumn{1}{c}{CIFAR10 (8)} & \multicolumn{1}{c}{Digit5 (4)}&\multicolumn{1}{c}{Office (4)} &\multicolumn{1}{c}{Domainnet (5)} \\
    \hline
    \hline
    \textsc{Local} & 79.66$\pm$0.55 & 40.90$\pm$0.92 & 40.65$\pm$0.84 & 83.09$\pm$0.42 & 60.61$\pm$0.78 & 48.33$\pm$0.27\\
    \hline
    \textsc{FedAvg} & 82.97$\pm$0.67 & 47.23$\pm$0.25 & 49.61$\pm$0.65 & 83.73$\pm$0.55 & 62.76$\pm$0.76 & 49.11$\pm$0.47\\
    \textsc{FedProx} & 83.61$\pm$0.64 & \textbf{49.19}$\pm$0.89 & \textbf{50.76}$\pm$0.26 & 83.92$\pm$0.85 & 62.99$\pm$1.07 & 49.32$\pm$0.53\\
    \textsc{FedDF} & 83.20$\pm$0.57 & 47.88$\pm$0.62 & 50.43$\pm$0.49 & 84.48$\pm$0.28 & * & 49.21$\pm$0.42 \\
    \hline
    \textsc{FedSplit} & 81.95$\pm$0.43 & 44.67$\pm$0.64 & 46.00$\pm$0.78 & 82.96$\pm$0.42 & 62.71$\pm$0.88 & 48.61$\pm$0.49\\
    
    \textsc{FedGen} & 81.66$\pm$0.46 & 44.98$\pm$0.49 & 45.57$\pm$0.59 & 82.55$\pm$0.65 & 62.70$\pm$1.05 & 47.86$\pm$0.64 \\
    \hline
    \hline
    \textsc{FedCG} (ours) & \textbf{83.81}$\pm$0.28 & 47.52$\pm$0.68 & 49.15$\pm$0.48 & \textbf{84.82}$\pm$0.40 & \textbf{67.34}$\pm$0.83 & \textbf{49.90}$\pm$0.18 \\
      \hline
    \end{tabular}
    \caption{Comparison of \textsc{FedCG} with baselines in terms of top-1 test accuracy. Results reported in \textbf{bold} are the best performance. * indicates no results is measured. The number in the parentheses indicates the number of clients.}
         \label{performance_eval}
\end{table*}


\begin{figure*}[ht!]
    \centering
       \subfigure[FMNIST]{\includegraphics[width=5.6cm,height=4.5cm]{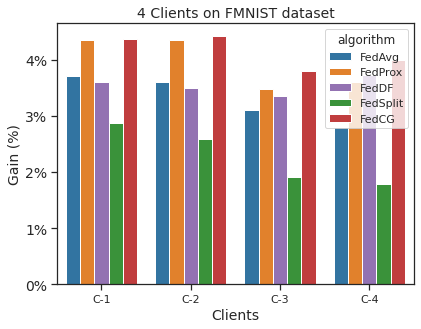} \label{fmnist_4}}
    \subfigure[CIFAR10]{\includegraphics[width=11.2cm,height=4.5cm]{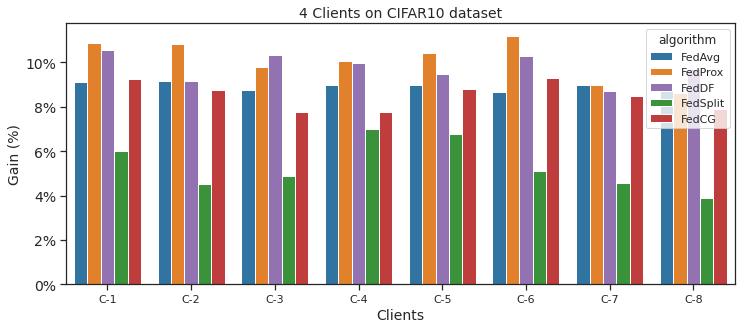} \label{cifar10_8}}
    \subfigure[Digit5]{\includegraphics[width=5.6cm,height=4.5cm]{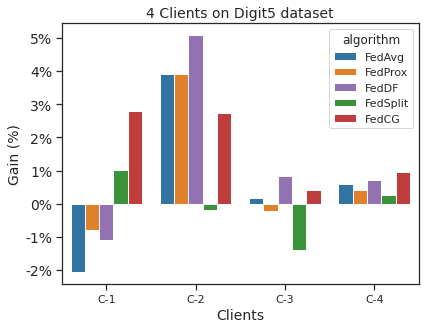} \label{digit_4}}
    \subfigure[Office]{\includegraphics[width=5.6cm,height=4.5cm]{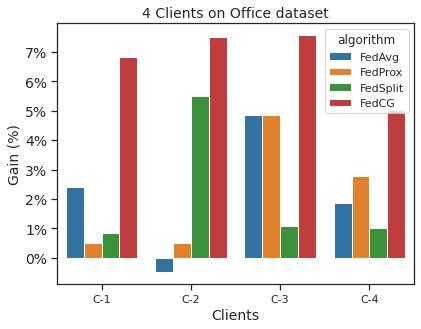}
    \label{office_4}}
    \subfigure[DomainNet]{\includegraphics[width=5.6cm,height=4.5cm]{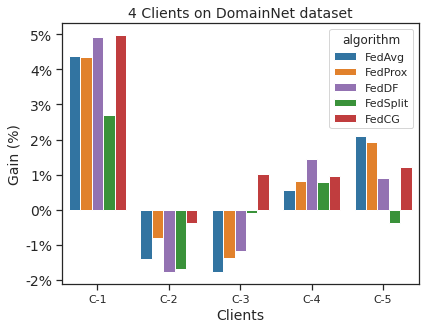}
    \label{domainnet_5}}
    \caption{Accuracy gains achieved by \textsc{FedAvg}, \textsc{FedProx}, \textsc{FedDF}, \textsc{FedSplit}, and \textsc{FedCG} (red) over \textsc{local} of each client on all 5 datasets. The vertical axis is the performance difference in terms of accuracy (\%). A positive (negative) gain means FL methods achieves better (worse) than the \textsc{Local} model.}
    \label{client_performance}
\end{figure*}

\subsubsection{Performance Evaluation}

We evaluate \textsc{FedCG}'s performance by first comparing its averaged clients' accuracy with those of 6 baselines in 6 scenarios. Table \ref{performance_eval} reports the results. It shows that \textsc{FedCG} achieves the best accuracy in 4 out of 6 scenarios, demonstrating its competitive performance. More specifically, \textsc{FedCG} achieves the best accuracy in all 3 non-IID scenarios. In particular, it outperforms the-next-best-performing \textsc{FedProx} by 4.35\% on Office. In IID scenarios, \textsc{FedAvg}, \textsc{FedProx} and \textsc{FedDF} have an edge in that they aggregate full local networks trained directly from original local data while there is no negative transfer effect caused by data heterogeneity~\cite{wang2019negative}. As a result, they perform better than \textsc{FedCG} overall, on CIFAR10 particularly. However, they are vulnerable to DLG attacks, as we will discuss in the next section. 

Because the objective of \textsc{FedCG} is to improve the performance of each client's personalized local network validating on local test data, we further compare the accuracy gains between \textsc{FedCG} and 5 FL baselines over the \textsc{Local}. Figure \ref{client_performance} shows the results. In IID scenarios, all FL methods outperform \textsc{Local} on all clients by large margins, as shown in Figures \ref{fmnist_4} and \ref{cifar10_8}. Particularly, \textsc{FedCG} performs best on FMNIST(4) while \textsc{FedProx} performs best on CIFAR10(8). In non-IID scenarios, while no FL method can beat \textsc{Local} on every client across all 3 non-IID datasets, \textsc{FedCG} achieves the best result such that it outperforms \textsc{Local} on 12 out of 13 clients, as shown in Figure \ref{digit_4}, \ref{office_4} and \ref{domainnet_5}. \textsc{FedAvg}, \textsc{FedProx} and \textsc{FedDF} are not excel in non-IID scenarios as they are in IID scenarios because the average-based global model may be far from client's local optima~\cite{li2021federated}. Besides, the distillation dataset leveraged by \textsc{FedDF} is from a different domain than those of clients, which may have negative impacts on the performance of aggregated global model. 



\subsubsection{Privacy Evaluation.}

\begin{table*}[ht!]
	\centering
	\begin{tabular}{c|c||c|c|c|c|c}
		\hline
		Dataset & Metric & \textsc{FedAvg} & \textsc{FedAvg} (0.001) & \textsc{FedAvg} (0.1) & \textsc{FedSplit} & \textsc{FedCG} \\
		\hline
		\hline
		\multirow{2}*{CIFAR10} & Accuracy(\%) & $47.23 $ & $37.5 $ & $35.4$ & $44.67$ & \textbf{47.52} \\
		
		& PSNR(dB) & $24.14$ & $20.30 $ & $6.81 $ & $6.23$ & \textbf{8.99}\\ 
        \hline
        \multirow{2}*{DIGIT} & Accuracy(\%) & $83.73 $ & $68.6 $ & $64.3 $ & $82.96 $ & \textbf{84.82}  \\
	
		& PSNR(dB) & $26.82 $ & $20.20$ & $6.32$ & $5.47 $ & \textbf{7.85}\\
		\hline
		\multirow{2}*{OFFICE} & Accuracy(\%) & $62.76 $ & $40.6 $ & $32.7$ & $62.71$ & \textbf{67.34}  \\
		
& PSNR(dB) & $23.14 $ & $18.78$ & $6.38 $ & $5.61 $ & \textbf{7.57} \\
		\hline
	\end{tabular}
	\caption{Comparison of \textsc{FedAvg}, \textsc{FedSplit} and \textsc{FedCG} in terms of model performance and privacy-preserving capability. The numerical number in the parentheses indicates the noise level $\sigma^2$. }
\label{table:dp}
\end{table*}

\begin{figure}[ht!]
  \centering
  \includegraphics[width=.45\textwidth]{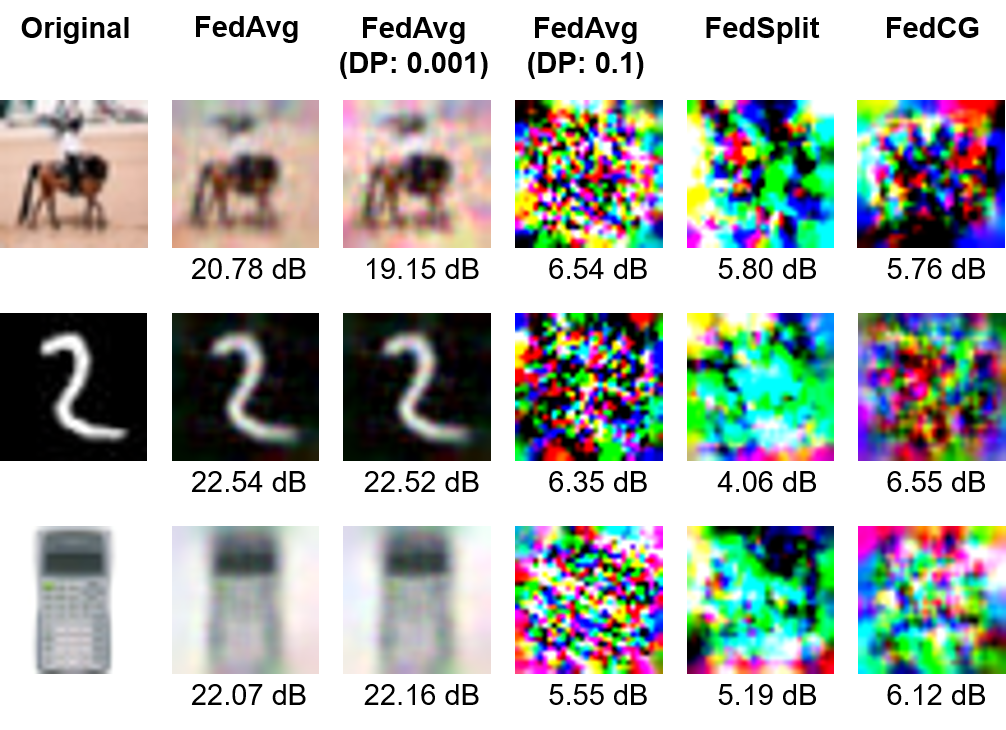}
  \caption{Reconstructed images using DLG attack in \textsc{FedAvg},  \textsc{FedSplit} and \textsc{FedCG}. From the first row to the last row, images are taken from CIFAR10, DIGIT5 and OFFICE respectively. PSNR is reported under each restored images. }
  \label{fig:reconstructed_images}
\end{figure}

We calculate Peak Signal-to-Noise Ratio (\textsc{PSNR}) to measure the similarity between original images and images recovered from DLG. PSNR is an objective standard for image evaluation, and it is defined as the logarithm of the ratio of the squared maximum value of RGB image fluctuation over MSE between two images. The higher the PSNR score, the higher the similarity between the two images. We also apply differential privacy (DP) to \textsc{FedAvg} by adding Gaussian noises to shared gradients. We experiment with two noise levels, $\sigma^2=0.1$ and $\sigma^2=0.001$. 

Table \ref{table:dp} compares \textsc{FedCG} with \textsc{Fedsplit} and \textsc{FedAvg} in terms of model performance measured by accuracy and privacy-preserving capability measured by PSNR between the ground-truth image and the recovered one using DLG. Compared to \textsc{FedAvg}, \textsc{Fedsplit} and \textsc{FedCG} could achieve competitive accuracies for all three datasets. However, native \textsc{FedAvg} has much higher risk of leaking data information (high PSNR value), which can also be observed in Figure \ref{fig:reconstructed_images} where the reconstructed images from \textsc{FedAvg} looks very similar to the originals. Although the data privacy could be better protected by introducing DP to \textsc{FedAvg}, there is a significant drop (over $20\%$) in the model performance. On the other hand, \textsc{FedSplit} and \textsc{FedCG} effectively protect the data privacy (low PSNR value). This is also evidenced by the reconstructed images in Figure \ref{fig:reconstructed_images}, where no clear patterns of the reconstructed images could be observed.  While achieving similar privacy protection level, \textsc{FedCG} demonstrates better model accuracy than \textsc{FedSplit} for all three datasets by at least $2\%$. 

\section{Conclusion}
We propose \textsc{FedCG}, a novel federated learning method that leverages conditional GAN to protect data privacy while maintaining competitive model performance. \textsc{FedCG} decomposes each client's local network into a private extractor and a public classifier, and keeps the extractor local to protect privacy. It shares clients' generators with the server to aggregate shared knowledge aiming to enhance the performance of clients' local networks.  Experiments show that \textsc{FedCG} has a high-level privacy-preserving capability and can achieve competitive model performance. Future works include extending \textsc{FedCG} to deeper neural networks for investigating the effectiveness of \textsc{FedCG} further and providing theoretical analysis on the privacy guarantee of \textsc{FedCG}.


\bibliography{aaai22}

\clearpage
\onecolumn

\twocolumn

\end{document}